\documentclass[letterpaper, 10 pt, conference]{ieeeconf}  % Comment this line out if you need a4paper

\IEEEoverridecommandlockouts                              % This command is only needed if 
\usepackage{graphics} % for pdf, bitmapped graphics files
\usepackage{amsmath} % assumes amsmath package installed
\usepackage{amssymb}  % assumes amsmath package installed
\usepackage{comment}
\usepackage{url}
\usepackage{graphicx}
\usepackage{subcaption}
\usepackage{xcolor}
\usepackage[ruled,vlined]{algorithm2e}
\usepackage[colorinlistoftodos]{todonotes}

\usepackage{xcolor}

\newcommand{\emre}[1] { {#1} }

\title{\LARGE \bf
Reward Conditioned Neural Movement Primitives for Population Based Variational Policy Optimization
}

\author{M.Tuluhan Akbulut$^\dag$, Utku Bozdogan$^\dag$, Ahmet Tekden$^\dag$ and Emre Ugur$^\dag$% <-this % stops a space
\thanks{*This work was supported by the European Union’s Horizon2020 research and innovation programme under grant agreement no. 731761, IMAGINE.}
\thanks{$^\dag$ Authors are with the Department of Computer Engineering, Bogazici University, Turkey
        {\tt\small tuluhan.akbulut@boun.edu.tr}}%
}

\begin{document}

\maketitle
\thispagestyle{empty}
\pagestyle{empty}

%%%%%%%%%%%%%%%%%%%%%%%%%%%%%%%%%%%%%%%%%%%%%%%%%%%%%%%%%%%%%%%%%%%%%%%%%%%%%%%%
\begin{abstract}

The aim of this paper is to study the reward based policy exploration problem in a supervised learning approach and enable robots to form complex movement trajectories in challenging reward settings and search spaces. For this, the experience of the robot, which can be bootstrapped from demonstrated trajectories, is used to train a novel Neural Processes-based deep network that samples from its latent space and generates the  required trajectories given desired rewards. Our framework can generate progressively improved trajectories by sampling them from high reward landscapes, increasing the reward gradually. Variational inference is used to create a stochastic latent space to sample varying trajectories in generating population of trajectories given target rewards. We benefit from Evolutionary Strategies and propose a novel crossover operation, which is applied in the self-organized latent space of the individual policies, allowing blending of the individuals that might address different factors in the reward function. Using a number of tasks that require sequential reaching to multiple points or passing through gaps between objects, we showed that our method provides stable learning progress and significant sample efficiency compared to a number of state-of-the-art robotic reinforcement learning methods. Finally, we show the real-world suitability of our method through real robot execution involving obstacle avoidance.
\end{abstract}

%%%%%%%%%%%%%%%%%%%%%%%%%%%%%%%%%%%%%%%%%%%%%%%%%%%%%%%%%%%%%%%%%%%%%%%%%%%%%%%%
\section{INTRODUCTION}
\emre{In the last decade, robot learning has become a key technology for equipping robots with dexterous robot skills. Acquiring robotic skills requires learning of multi-modal sensorimotor relations in connection with external parameters and goals. An effective approach to equip the robots with advanced skills is to first teach the desired skill via Learning from Demonstration (LfD), and then enable the robot to improve its skill by self-exploring the world and task space via reinforcement learning (RL) \cite{kroemer2019review,kober2013reinforcement,Argall2009}. LfD is generally framed as a supervised learning (SL) problem where the target sensorimotor trajectories are provided by an expert and the model parameters are tuned to produce these trajectories. On the other hand, the target trajectories are not directly provided in RL, and the robot updates its policy parameters to obtain maximal reward through trial-and-error exploration. Sample efficiency is critical for RL when applied in the real-world since experimenting with the real robot takes time, and resources. Therefore, LfD approaches are widely used to bootstrap this search, effectively giving a prior to constrain policy search so that a solution can be found faster and safer \cite{dulac2019challenges,hester2018deep, vecerik2019practical,rajeswaran2018learning}.

On the relationship of RL with SL, Barto and Dietterich argued that solving RL problems using SL is not possible because SL is supposed to use an error between desired and produced output however RL settings only provide evaluation signals based on the performance of the produced output and do not provide direct knowledge about desired output \cite{barto2002}. Recently, Schmidhuber \cite{schmidhuber2020reinforcement} proposed to use the environmental feedback (reward) as input together with time horizon with Upside Down RL (UDRL) method to close the gap between RL and SL. \cite{srivastava2019training} realized this method in Atari environments and showed that UDRL outperforms traditional RL methods. On the other hand, robotic settings have further challenges and requirements. While this method was shown to predict discrete actions given the current state and the desired future rewards, robotic tasks require the generation of complex continuous sensorimotor trajectories. Moreover, rather than receiving reward from state-action pairs, robotic systems typically receive reward for the complete trajectory after executing the complete action. 

In this paper\footnote{The source code and data are provided to the community \url{https://github.com/mtuluhanakbulut/RC-NMP}}, we also frame robotic reward-based learning as a supervised learning problem. In order to address the challenges listed above, we propose a reward-conditioned neural policy architecture that is built on top of Neural Processes \cite{NP} that can encode multi-modal distributions in relation with non-linearly related external parameters in a robust and effective way. After trained completely via SL, given the desired reward and other relevant parameters, our method exploits the formed robust latent representation, and generates the complete motion trajectory in the corresponding robotic task.

We benefit from Evolutionary Strategies (ES) \cite{nolfi2000evolutionary} and variational inference \cite{Blei_2017} in order to effectively explore the challenging search space of the robotic problems that potentially include multiple solutions and multiple local minima difficult to escape. For this, we propose a novel crossover operation, which is applied in the self-organized latent space of the individual policies, allowing the temporal blending of the trajectories of the corresponding individuals that might solve different sub-goals or address different factors in the reward function. 
%We observed that this operation accelerates the learning, because sub-parts of undesired trajectories, which receive rewards from different portions, could possibly be combined to achieve high rewards at early steps. 
We also use a mutation operation to increase the diversity and make small movements in the search space by applying Gaussian noise in the task space. 
%This operation come in useful to make small adjustments to trajectory distribution. \question{Belki bu silinebilir son deneyde smooth Gaussian koymak da iyi iş yapıyordu.}
Finally, variational inference is used to create a stochastic latent space to sample varying trajectories conditioned on rewards.

In summary, the contributions of this paper are as follows:
\begin{itemize}
    \item A novel reward conditioned neural movement primitive, which generates trajectories by sampling from its latent space conditioned on rewards, is proposed and implemented. Our framework can generate progressively improved trajectories by sampling them from high reward landscapes, increasing the reward gradually. 
    \item Variational inference in the latent space of movement primitive is realized. The trajectory generation capability is compared against Conditional Neural Movement Primitives (CNMP) \cite{Seker2019} that does deterministic sampling, and our system is shown to be effective in producing multiple trajectories that cover the provided demonstrations.
    \item A novel crossover operator in the self-formed policy latent space is proposed, and its effectiveness in bootstrapping learning of complex trajectories is validated quantitatively. 
    \item A complete learning and execution cycle is implemented and its performance is compared against several state-of-the-art RL methods \cite{Akbulut2020,peters2010relative,colome2017dual}. Our method is verified in a number of simulation tasks including ones that provide sparse rewards, and verified through real-robot execution that requires generation of complex trajectories. It is shown to have a more stable learning progress compared to \cite{Akbulut2020} and provide significant sample efficiency compared to \cite{peters2010relative,colome2017dual}.
\end{itemize}
}

\section{RELATED WORK}

\paragraph*{Learning Movement Primitives}

\emre{
Bootstrapping the movement skills via Learning from Demonstration (LfD) has been effectively used in robot learning problems \cite{Argall2009,kroemer2019review}. Learning frameworks that are based on dynamic systems \cite{ijspeert2013dynamical}, probabilistic modeling \cite{Calinon2016} and combination of them \cite{Girgin2018,Ugur2020} have been popular in recent years. Dynamic Movement Primitives (DMP) learned the demonstrated trajectory as a set of differential equations, extending a spring-mass-damper system with a non-linear function. While DMP generated deterministic trajectories, Probabilistic Movement Primitives (ProMP) \cite{Paraschos2013} can encode a distribution of trajectories and generates stochastic policies.
Conditional Neural Movement Primitives (CNMP) \cite{Seker2019}, built on Conditional Neural Processes (CNP) \cite{CNP}, can also encode trajectory distributions, and it can additionally learn non-linear relationships between task parameters and complex trajectories from few data. Our method is built on a similar neural network model, namely Neural Processes (NP) \cite{NP}. Unlike CNMPs, our network is conditioned with desired rewards and uses stochastic sampling in the latent space.
}

\paragraph*{Variational Inference in Movement Primitives}
\emre{
Variational inference has started being used to produce movement primitives in recent years \cite{Chen2016,Noseworthy2020,Osa_2020}. \cite{Chen2016} used Deep Variational Bayes Filtering (DVBF) \cite{karl2017deep} to embed DMPs in Variational Auto Encoders (VAE). \cite{Noseworthy2020} used Temporal Convolution Networks \cite{oord2016wavenet} to model trajectories and trained VAE conditioned on task parameters. \cite{Osa_2020} used VAE with continuous and discrete latent variables and conditioned the decoder network on goal parameters. All these methods require training trajectories in a scale of thousands. On the other hand, our model can learn from a small number of training trajectories because the underlying neural network model \cite{NP} uses random observation samples to encode trajectories. 
%Our approach is based on Neural Processes(NP) \cite{NP} like CNMPs are based on Conditional Neural Processes \cite{CNP} with a difference that we are using a Gaussian prior with zero mean and unit variance for the latent distribution. 
}

\paragraph*{Adaptation of Primitives}
\emre{Encountered with novel environments, tasks or situations, the  underlying parameters of the movement primitives or policies are adjusted via trial-and-error and reward based reinforcement learning (RL) \cite{kober2013reinforcement}.
RL is often bootstrapped by LfD to facilitate sample-efficient learning \cite{dulac2019challenges}. In this approach, expert demonstrations are used to limit the large search space in which RL looks for a solution to manageable volumes by providing  well performing initial experiences to the algorithm \cite{hester2018deep, vecerik2019practical,rajeswaran2018learning}.
\cite{Stark2019,Ewerton2019} also combined LfD and RL in Movement Primitives framework, by exploiting ProMPs to encode demonstrations and adapt to new task constraints. In \cite{Stark2019}, separate ProMP models learned the task of pushing an object to different target positions from demonstrations. This skill is extended to a new target position with a new ProMP using Relative Entropy Search (REPS)  \cite{peters2010relative}, where KL divergence is used to preserve shapes. \cite{Ewerton2019} combined ProMPs and Gaussian Processes to condition ProMPs with task parameters. RL is used to learn the relations between the environment parameters and parameters of the model, exploiting a trajectory relevance metric. Adaptive Conditional Neural Movement Primitives (ACNMP) \cite{Akbulut2020} can also encode different task parameters in a single model providing order-of-magnitude sample efficiency compared to \cite{Stark2019,Ewerton2019}. Contrary to the previous models, ACNMP does not require explicit optimization using metrics such as relevance or KL-divergence. Instead, ACNMP is trained together with the demonstrated trajectories and the newly explored ones, automatically preserving the old skills while extending the model to the new task parameters thanks to the robust and flexible representations generated. In this paper, we compare the performance of our population-based approach with ACNMP in terms of learning stability and performance.

\begin{figure*}[t]
    \centering
    \includegraphics[width=0.92\linewidth]{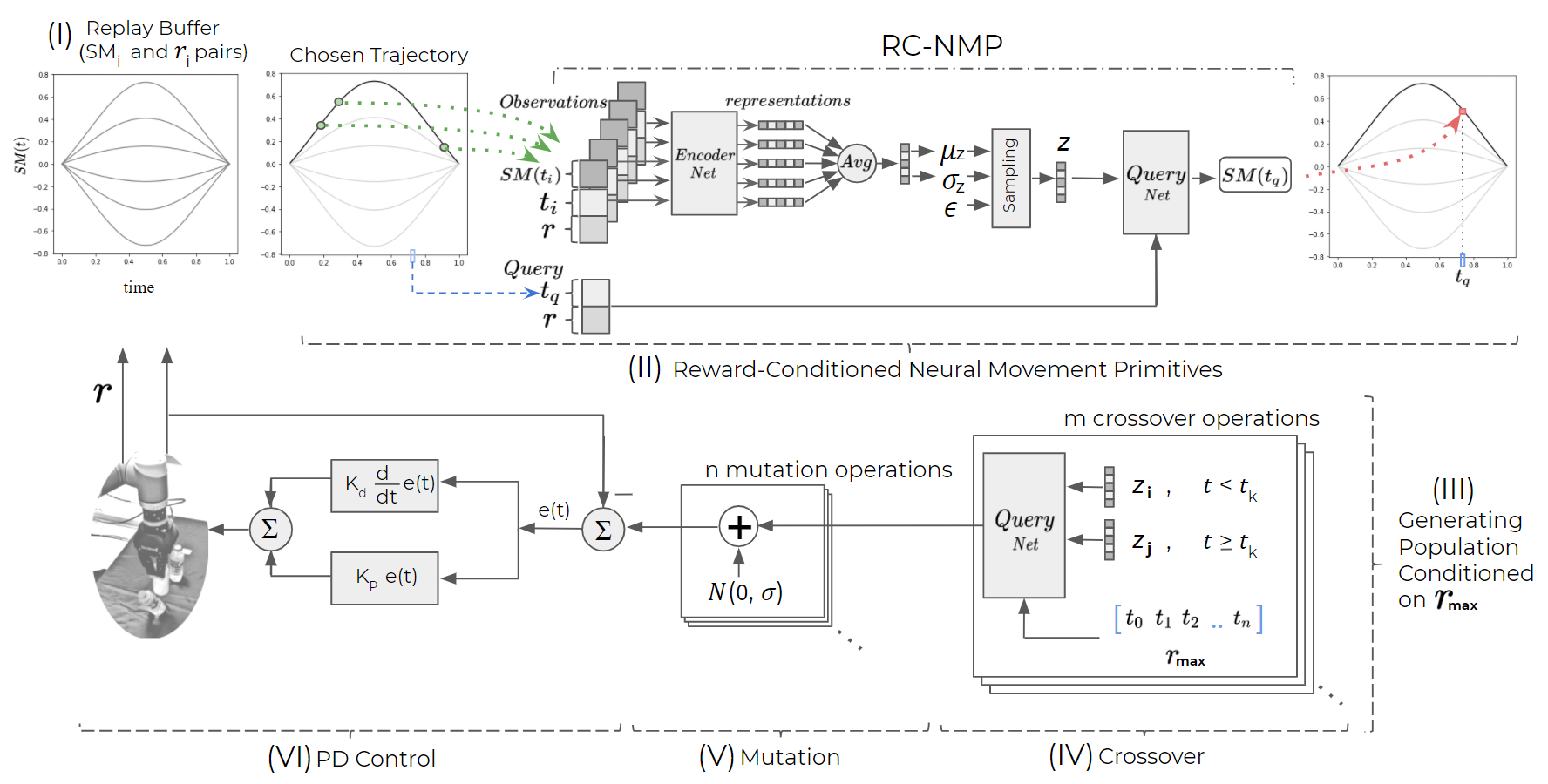}
    \caption{ Training the RC-NMP by sampling observations from the demonstration set and predicting the trajectory value over the queried time-step and generating a trajectory according to best reward parameter at the test time. %See Section \ref{Method}.A for more details.
    }
    \vspace{-0.3cm}
    \label{g}
\end{figure*}
\paragraph*{Evolutionary policy optimization}
Genetic algorithms have recently gained popularity as their optimization procedures can effectively deal with challenging search spaces with multiple local minima. They also scale well in some problems compared to RL \cite{salimans2017evolution,such2017deep}.  Genetic-Gated Networks \cite{chang2018genetic} used a chromosome vector, which is subject to mutation and crossover, controlling the gating choices of the policy network. Genetic algorithm is used for optimization of this vector. 
The chromosome vector representing the best choice is taken and the resulting network is subjected to gradient based optimization. Genetic Policy Optimization \cite{gangwani2017policy} utilized both imitation learning and deep reinforcement learning within. Two parent policy networks are subjected to crossover in the state space via applying imitation learning on their offspring in order to generate a state space distribution similar to its parents'. Mutation operation is modelled by policy gradient methods, and updated with high variance gradients. Recently, Natural ES is applied directly on network parameters \cite{wang2020instance}. The learning problem is tackled for dynamic environments unlike previously mentioned approaches and the process is regulated by instance weighting which assigns higher weights to instances based on their novelty and quality. \cite{conti2018improving} showed that their ES approach performed better on dynamic tasks designed to be deceptive compared to certain baseline RL, ES and meta learning approaches, since it encouraged different behaviours more through novelty search \cite{lehman2011novelty}. \cite{khadka2018evolution,bodnar2019proximal,pourchot2018cem} combined ES and Deep Deterministic Policy Gradient algorithms. In our work, unlike the previous ones, a novel crossover operation that allows temporal blending of the parents is applied in the self-organized latent space of the policies, and the individuals are generated in a dynamic process by our reward conditioning mechanism.

\paragraph*{Use of learning data with varying quality}
Our approach encourages exploration in more rewarding states by conditioning on higher rewards, similar to a directed exploration. On the other hand, in order to improve exploration, we use all experiences of the robot independent of the quality of the instances. In Double REPS \cite{colome2017dual} the problem of learning from bad experiences is specifically addressed. Their approach improved upon REPS \cite{peters2010relative} by adding conditions to constrain policy updates such that the policy stayed close to high reward samples while also explicitly trying to stay away from low reward samples. This is done via clustering of samples based on their features and transformed rewards for each rollout, which acted as attractive and repulsive fields, respectively. In our paper, we compare the performance of our approach with REPS and DREPS in terms of sample efficiency and performance.} 

\section{PROPOSED METHOD}
\label{Method}
\emre{We propose a framework that follows a learning and policy update loop shown in Fig.~\ref{g}, where the experience of the robot stored in Replay Buffer (I) is used to train a deep network that generates the required trajectories given desired rewards (II). Then, a population of trajectories is generated by this network to obtain maximal reward (III), and evolutionary operations are applied to this population intermixing the representations and increasing diversity of the solutions (IV, V). Finally, the generated trajectories are executed in the given environment and task, and added into the Replay Buffer with the observed reward value (VI). In the rest of this section, the architecture and proposed operations are provided in detail.

\paragraph{Replay Buffer (I)} The method starts from an initially demonstrated sensorimotor (SM) trajectory or a random one and learns to generate motion trajectories that maximize the reward function. The Replay Buffer stores the experienced SM trajectories and the corresponding rewards. Next, using the set of trajectories and reward values from the Replay Buffer, our system learns a model that generates SM trajectories given target reward values as follows.

\paragraph{Reward Conditioned Neural Movement Primitives (RC-NMP) (II)}\footnote{We are interested in problems where there is a reward function ($R$) that maps trajectories ($\tau$)  to real valued rewards: $R(\tau) = r , r \in \mathbb{R}$. Standard reinforcement learning algorithms maximize expected reward $\mathbb{E}_{\tau\sim p_\theta(\tau)}[R(\tau)]$ with respect to parameters $\theta$. Our model, on the other hand, is designed to generate the trajectory distribution that gives the global maximum or a satisfactory local maximum ($r^*$) as reward. To benefit from supervised learning, we take a more direct approach by conditioning the trajectory distributions w.r.t. rewards $r$ and our model optimizes its parameters to find the trajectory distribution conditioned on ($r^*$).} 
Fig.~\ref{g} (II) shows how our model is trained using (SM, r) pairs. Our RC-NMP model has a specific deep encoder-decoder neural network architecture built on top of Neural Processes \cite{NP}. The encoder layer of the RC-NMP learns a representation of trajectory points ($z$) conditioned on the time and the corresponding reward. The decoder layer takes the learned representations with the reward and outputs the trajectory as a function of time. Fig.~\ref{g} (II) shows the training procedure using a hypothetical 1D scenario. At each training iteration, random sensorimotor time and value pairs (the green dots in the figure), named observation points ($O$), are sampled from a trajectory randomly chosen from the Replay Buffer. These (SM, t) points are processed by a parameter sharing encoder network and transformed into their corresponding latent space representations, and then merged into a common latent representation with the averaging operation. This latent representation space is modelled as a Gaussian distribution $N(0, I)$ and is optimized using variational inference similar to $\beta$-Variational Auto-Encoders ($\beta$-VAE) \cite{Higgins2017} and Conditional Variational Auto-Encoders (CVAE) \cite{Sohn2015}. The log-likelihood of a trajectory is composed of the log-likelihoods of individual data points: $\log p_\theta(\tau|t,r) = \sum\limits_{t=1}^T \log p_\theta(x_t|t,r)$.
Using the derivation in \cite{Diederik2019}, this log-likelihood can be expressed as:
\begin{equation}
\begin{aligned}
\log p_\theta(x_t|t,r) \quad & = \mathbb{E}_{q_\phi(z|O,r)}[\log p_\theta(x_t|t,r)] \\
\quad & = \mathbb{E}_{q_\phi(z|O,r)}[\log \frac{ p_\theta(x_t,z|t,r)}{q_\phi(z|O,r)}] \\
\quad & + \mathbb{E}_{q_\phi(z|O,r)}[\log \frac{ q_\phi(z|O,r)}{p_\theta(z|x_t,t,r)}]
\end{aligned}
\end{equation}

The first term before the plus is the evidence lower bound (ELBO) and the second term is a KL divergence term, which is non-negative. Therefore, ELBO is a lower bound on the likelihood of data. By following steps in \cite{Higgins2017}, ELBO can also be expressed as:
\begin{equation}
\label{objective}
\begin{aligned}
\mathcal{L}(\theta,\phi) \quad & = \mathbb{E}_{q_\phi(z|O,r)}[\log p_\theta(x_t|z,t,r)] \\
\quad & - \beta D_{KL} \left(q_\phi(z|O,r) \middle\| p_\theta(z|t,r) \right)
\end{aligned}
\end{equation}

The prior $p_\theta(z|t,r)$ can be replaced with a Gaussian $N(0, I)$ \cite{doersch2016tutorial}. Our proposed model is different from CNMPs\cite{Seker2019}, in the sense that the network is specifically designed to be conditioned with target rewards. Furthermore, the learned latent representations in our case are represented stochastically contrary to deterministic representation formation in CNMPs. This change enables the model to represent the trajectory distribution in latent space rather than in the task space.

\paragraph{Generating population of individual solutions (III)} After training the RC-NMP with the Replay Buffer experience, our system undergoes a search for better solutions, by sampling trajectories in the vicinity of the best trajectories. For this, multiple solutions are generated by conditioning the network with the maximum reward obtained until that point. Stochastic sampling procedure generates different latent representations, $z_i$. In order to further maximize the diversity of these solutions, two evolutionary operations are applied as follows.

\paragraph{Crossover operation (IV)} The decoder network, i.e. Query Net, receives the latent representation, and target reward and time-points in order to generate the trajectory points. In order to intermix the representations of different individuals generated in the previous step, we applied one-point crossover through temporal blending on the corresponding latent representations. For this, random pairs of latent representations $(z_i,z_j)$ are selected for crossover; and a random time-point $t_k$ is selected from the $t_0-t_n$ range. Then the query network is conditioned with $z_i$ for time-points $t_0-t_k$ and with $z_j$ for time-points $t_{k+1}-t_n$; generating SM values for all time-points. This crossover operation is repeated for $m$ such random pairs, generating $m/2$ trajectories. Applying crossover symmetrically on the selected pairs, $m$ total trajectories are obtained at the end of the crossover operation.

\paragraph{Mutation operation (V)}
In order to increase the diversity of the solutions and encourage further exploration, a smoothed Gaussian noise is added to all $n$ trajectories to encourage further exploration in the trajectory space. Note that as the underlying network model does not have extrapolation capability in the latent space, big changes in the latent space resulted in slight changes in the trajectory space, therefore mutation operation is applied in the task space rather than the latent space.

\paragraph{PD controller (VI)}
We incorporated a PD controller in our model to ensure smooth movement and guarantee of reaching to the goal. The trajectories obtained at the end of the evolutionary operations are given to PD controller for execution. Our controller has two objectives: following the trajectory smoothly and reaching to the goal point at the end. 
\begin{equation}
\begin{aligned}
u(t) \quad & = K_p*e(t) - K_d*\frac{de(t)}{dt} \\
e(t) \quad & = \lambda_1 * (g-x(t)) + \lambda_2 * (\tau(t)-x(t))
\end{aligned}
\end{equation}

Here $g$ denotes the goal position and $x(t)$ and $\tau(t)$ denote the current and desired positions at time $t$. $K_p$ and $K_d$ are PD parameters that define control behaviour. $\lambda_1$ and $\lambda_2$ are weight parameters of the two objectives. In our application, $\lambda_1$ grows exponentially and $\lambda_2$ decreases exponentially between 0 and 1 over time to ensure that the robot follows the trajectory as much as it can.
}

\emre{After the execution, the combination of the best performing trajectories and random ones are added to the Replay Buffer and the process is repeated.}

\section{EXPERIMENTS}
    
    \begin{figure}
    \centering
    \includegraphics[width=0.95\linewidth]{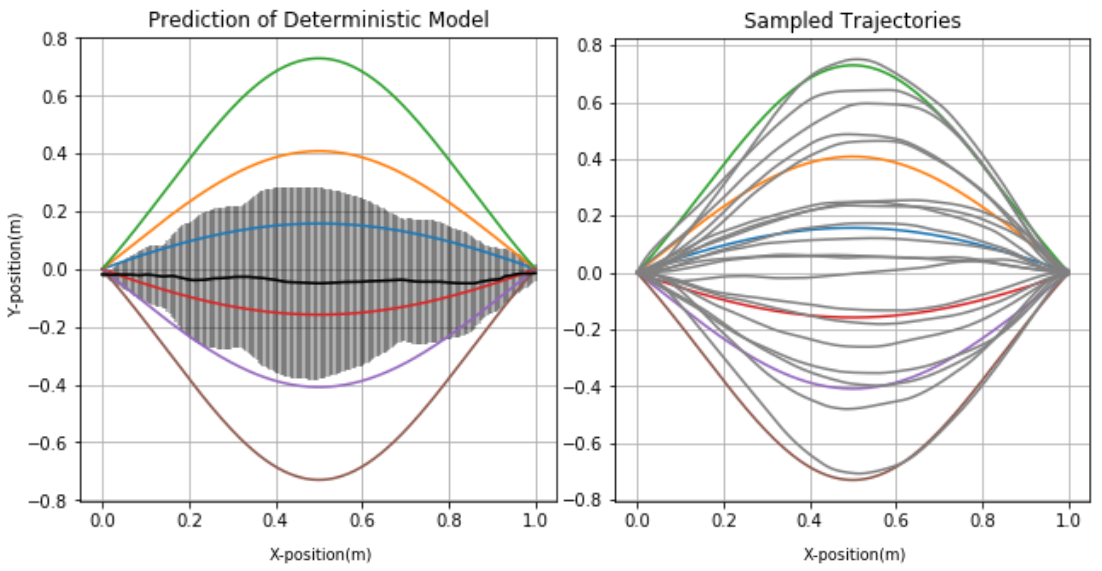}
    \caption{
    6 demonstrations are shown with colored curves in both figures. Given only the initial position, the generated trajectories for CNMP (left) and RC-NMP (right) are compared.\vspace{-0.5cm}}
    \label{fig:exp1}
\end{figure}
Experiments are conducted to evaluate RC-NMP compared to the state of the art methods mentioned in the paper. Details can be found in the readme file of the source code.

\subsection{Stochastic sampling in latent space}

\emre{
This experiment aims to analyze the trajectories generated by our proposed stochastic latent space sampling method and compare these trajectories with the trajectory generated by the CNMPs \cite{Seker2019}. For this, 6 demonstration trajectories, provided with the colored lines in Fig.~\ref{fig:exp1}, are learned by our method and the CNMPs. These trajectories start from the same initial position, follow varying curved paths and ends up in the same final position. After learning, both models are conditioned with the initial position, i.e. models are requested to generate trajectories that start from the initial position. As provided in Fig.~\ref{fig:exp1} (left), based on the distribution shown with the shaded area, CNMP generated a single trajectory shown with the bold line. On the other hand, as shown in Fig.~\ref{fig:exp1} (right), our method was able to generate different trajectories that well-covered the demonstrated range from multiple runs due to our latent representation stochastic sampling approach. Note that our method was able to generate trajectories that were similar to the demonstrated trajectories at the edges, i.e. to the top-most and bottom-most ones.
}

\subsection{Performance in generating complex trajectories}

\begin{figure}[t]
    \centering
    \includegraphics[width=0.95\linewidth]{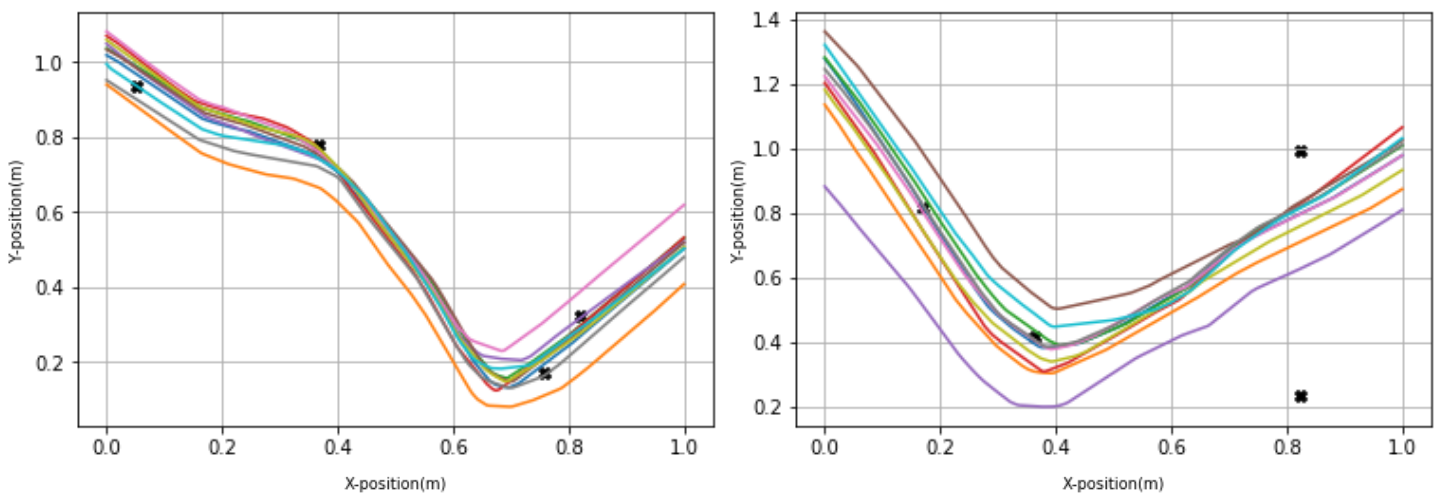}
    \caption{Example environments for the task generating complex trajectories. The complexity is dependent on the positions of random points and affects the model's performance. \vspace{-0.5cm}}
    \label{fig:exp2}
\end{figure}

\begin{figure}
    \includegraphics[width=\linewidth]{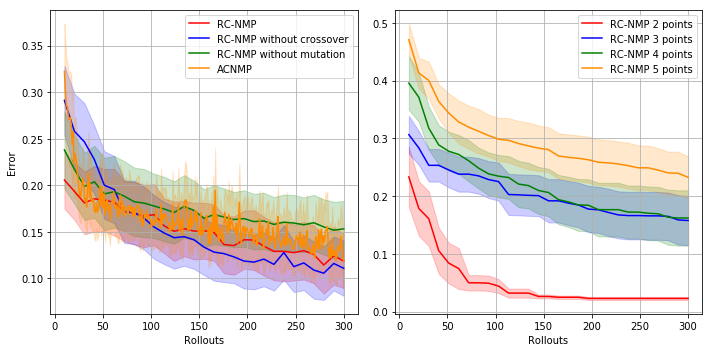}
    \caption{Learning curves for RC-NMP, RC-NMP w/o crossover, RC-NMP w/o mutation and ACNMP for generating complex trajectories experiment on the left, learning curves of RC-NMP for different number of random points on the right. \vspace{-0.5cm}}
    \label{fig:exp2-1} 
\end{figure}

\emre{In this experiment, the aim is to thoroughly analyze the performance of our method in generating target complex trajectories, and compare its performance with ACNMP \cite{Akbulut2020}. Furthermore, in order to understand the importance and influence of the crossover and mutation operations, we conducted ablation procedure where either crossover or mutation was removed from the learning loop. The task is designed as generating trajectories that pass through given target points. For each task, 2 to 5 different random points are generated from a uniform distribution. The goal of the models is to learn to generate trajectories that pass through each set of the generated target points. For this task, the reward function is defined as the minus sum of the distances of each target point to the generated trajectory. Two example tasks with 4 target points are provided in Fig.~\ref{fig:exp2} where the points are shown with black dots. Every set of random points contains 10 different environments. For each environment, the algorithms run until they use 300 trajectory samples. 
Fig.~\ref{fig:exp2-1} (left) provides the change in error throughout the learning trials obtained by the compared models. The bold lines and shades show the mean and standard error of the sum of distances divided by the number of points achieved by each model in a mixed set of environments. As shown, the crossover operation makes the algorithm progress faster at early stages because genetic diversity at the beginning is vast, although the trajectories have a high error. The well-performing parts of 2 sub-optimal trajectories can be combined with this method and higher rewards can be achieved within a few generations. We also observe that without mutation operation, which makes small changes in the trajectories, the progress stopped when the generated solution was close to an optimal one. RC-NMP, on the other hand, was able to generate trajectories with low error already in the initial steps of learning. ACNMP performs slightly worse compared to our method. But more importantly, ACNMP has an oscillating performance change whereas RC-NMP offers a more stable learning because it learns the mapping from the reward values to the trajectories considering both well and poor performing experience rather than taking gradient steps from bad trajectories to good trajectories for those particular samples.  Fig.~\ref{fig:exp2-1} (right) provides the error made by our system for a different number of target points. We observe that our method could find perfect trajectories that pass through 2 points but the performance degrades with the increasing number of points. The reason behind this fall in the performance is related to the complex and challenging configuration of the points. For example, Fig.~\ref{fig:exp2} provides two different sets of target points and the generated trajectories. On the left, it can be seen that the points were well-spaced along the x-axis and the model was able to generate trajectories that satisfied the task requirements. However, the last 2 points in the right figure required trajectories with sharp turns as they were far away from each other along the y-axis and very close along the x-axis; and our model was not able to find a trajectory that passed through these two points. } 


\begin{figure}[t] 
    \includegraphics[width=\linewidth]{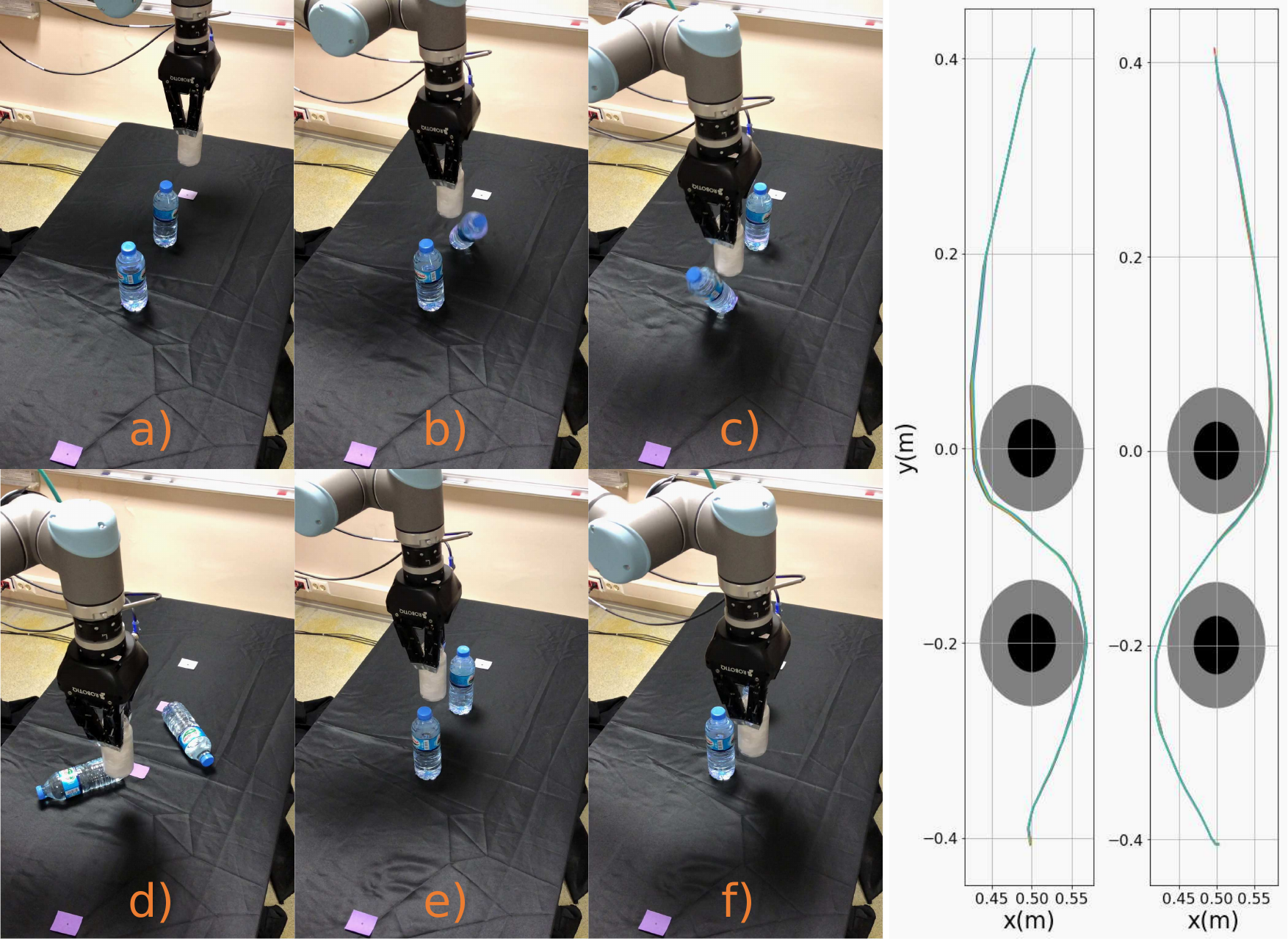}
    \caption{Snapshots from the `pass between two obstacles' experiment. (a) gives the setup; (b,c,d) show snapshots from collisions observed during learning and (e,f) show snapshots from successful execution after learning is accomplished. The generated trajectories for 2 modes are shown in the plots on the right. The black circle represents the bottle on the table and it is expanded with the radius of the bottle at hand to the gray circle. Video of this experiment is available in the supplementary material.\vspace{-0.6cm} }
    \label{fig:exp3}
\end{figure}

\subsection{Multi-modal problem: Passing through linearly aligned objects}
\label{experiment3}
\emre{In this section, we evaluate our method in a challenging planar obstacle avoidance task and compare the results with a state-of-the-art methods, REPS\cite{peters2010relative} and DREPS \cite{colome2017dual}, where this task was originally formalized. Initial position, two obstacles and final position are arranged in a line in this task. The end-effector of the robot needs to move from the initial position to the final position, avoiding the obstacles and passing through the aperture between them. The challenge in this task stems from the fact that this is a multi-modal problem with multiple distant optimal solutions where many approaches (e.g. \cite{peters2010relative}) get  trapped in between the alternative maxima as reported in \cite{colome2017dual}. Furthermore, the reward uses only a binary signal for obstacle collision, rather than a more informative value such as the  distance to the objects as in our previous experiments. In detail, we used the same reward defined in \cite{colome2017dual}. The reward is composed of three terms, where the first and second terms penalize collision with the bottles and not passing between the bottles, and the third term punishes long trajectories. Formally, the reward is defined as follows:
\begin{equation}
\textbf{R} = -2N_{\textrm{bottlesdown}} - 4\textbf{I}_{\textrm{cross}} - 0.15L_{\textrm{trajectory}} 
\end{equation}
where $N_{\textrm{bottlesdown}}$, $\textbf{I}_{\textrm{cross}}$, and $L_{\textrm{trajectory}}$ denote the number of bottles the robot collided with, whether the end-effector passed through the aperture between bottles, and the length of the trajectory in meters. Note that the second term does not enforce any constraint in the direction of aperture-pass. The models should generate an S-shaped or reverse S-shaped trajectory making the problem a multi-modal one.}

\begin{figure}
    \centering
    \includegraphics[height = 4.5cm]{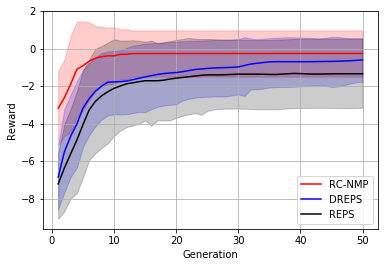}
    \caption{Learning curves for RC-NMP, DREPS and REPS for the multi-modal problem\vspace{-.6cm}}
    \label{fig:exp3-r}
\end{figure}

In the experiment, we solved the task in similar settings and conditions with \cite{colome2017dual} where the sum of radii of the bottle and the end effector was set as 6.5 cm. To reflect a similar setup with \cite{colome2017dual}, we put bottles with 3 cm radius on the table and another bottle with a radius of 3 cm was attached to the robot's gripper. We left 0.5 cm error margin as a rule of thumb since it is not possible to put the bottles at exactly the same locations between runs. The trajectories were run on UR10 robot.

%\emre{In the experiment, the sum of radii of the bottle and the end effector  was set as 6.5 cm. Similar to DREPS, we used simulated \question{simulated diyor muyuz, bunu tuluhan'a birakiyorum} experiences to learn the model and then transferred the end results to the real robotic system. We realized the experiment setup using UR10 robot. To reflect a similar setup with \cite{colome2017dual}, we put bottles with 3 cm radius on the table and another bottle with a radius of 3 cm was attached to the robot's gripper. We left 0.5 cm of error margin since it is not possible to precisely put bottles to the table on their exact locations.}

%\begin{figure*}[t]
%    \centering
%    \includegraphics[width=\linewidth]{figures/dreps.eps}
%    \caption{Snapshots from the experiment. After training, generated trajectories are demonstrated in real world setup. }
%    \label{fig:exp3}
%\end{figure*}

\emre{Our algorithm starts with a straight line in its replay buffer as in \cite{colome2017dual}. After the trajectories are generated, the PD controller is used to find the next position given the current and target positions in each time point. Our algorithm reached a satisfactory solution, reaching to the end position by passing through the bottles without colliding with them, 48 times out of 50 trials, whereas DREPS \cite{colome2017dual} had 47 times and REPS \cite{peters2010relative} had 35 times. Fig.~\ref{fig:exp3-r} provides a more detailed comparison between our method and the baselines. The bold lines and shades correspond to the mean and to standard deviation of rewards. As shown, our method (RC-NMP) reaches to maximum reward within 10 generations. Meanwhile, DREPS reaches to the same reward rate around 50 generations and REPS gets stuck at low rewards. Our method uses 20 trajectory samples in each generation and 20 trajectory samples from the crossover operation, as opposed to 50 trajectory samples used by DREPS and REPS per update. Therefore, our method required around 5 times less number of rollouts compared to the DREPS and REPS. Our model successfully generated trajectories for both modes and a number of snapshots from the execution of trajectories taken at different generations are provided in Fig.~\ref{fig:exp3} along with the top-down view of the final trajectory distribution.
}

\section{CONCLUSIONS}
%\question{Latent spaceteki jumpların farklılığa yol açmaması, mutationu fixed decoder weightlerinde perturbation olarak gerçekleştirmek, distributionun multi modeu aynı anda represente etmesini sağlamak}
In this work, we proposed a novel method, which synthetically produces its own demonstrations and improves itself fully using supervised learning. The model constructs a representation distribution using variational inference to sample reward conditioned trajectories and it increases the diversity of trajectories with the help of crossover and mutation, techniques from genetic algorithms. The experiments showed that it offers a more stable learning than its policy gradient variant and it is more sample efficient than two other RL methods implemented on top of movement primitives. In the future, we plan to investigate the means to achieve the mutation operation in the latent space and to keep the diversity in the population giving the capability of encoding multi-modal trajectories in the same policy.
% Learning from demonstration (LfD) often suffers from limited performance when extrapolation to other task parameters  and/or environments is required. The  applied remedy for this is usually the reward-based adaptation of the skills (i.e. RL) to account for the extrapolation asked. In this case, it becomes critical that the characteristics of the original demonstration are preserved while obtaining the desired novel skills. In this paper, we proposed a new adaptive movement primitive approach, namely ACNMPs, by integrating LfD and RL to address the challenges of efficient sampling, skill maintenance and skill transfer.

\addtolength{\textheight}{-2cm} 

%%%%%%%%%%%%%%%%%%%%%%%%%%%%%%%%%%%%%%%%%%%%%%%%%%%%%%%%%%%%%%%%%%%%%%%%%%%%%%%%

%%%%%%%%%%%%%%%%%%%%%%%%%%%%%%%%%%%%%%%%%%%%%%%%%%%%%%%%%%%%%%%%%%%%%%%%%%%%%%%%

%%%%%%%%%%%%%%%%%%%%%%%%%%%%%%%%%%%%%%%%%%%%%%%%%%%%%%%%%%%%%%%%%%%%%%%%%%%%%%%%
%\section*{APPENDIX}


%%%%%%%%%%%%%%%%%%%%%%%%%%%%%%%%%%%%%%%%%%%%%%%%%%%%%%%%%%%%%%%%%%%%%%%%%%%%%%%%
\bibliographystyle{IEEEtran}
\bibliography{references}

\begin{thebibliography}{10}
\providecommand{\url}[1]{#1}
\csname url@samestyle\endcsname
\providecommand{\newblock}{\relax}
\providecommand{\bibinfo}[2]{#2}
\providecommand{\BIBentrySTDinterwordspacing}{\spaceskip=0pt\relax}
\providecommand{\BIBentryALTinterwordstretchfactor}{4}
\providecommand{\BIBentryALTinterwordspacing}{\spaceskip=\fontdimen2\font plus
\BIBentryALTinterwordstretchfactor\fontdimen3\font minus
  \fontdimen4\font\relax}
\providecommand{\BIBforeignlanguage}[2]{{%
\expandafter\ifx\csname l@#1\endcsname\relax
\typeout{** WARNING: IEEEtran.bst: No hyphenation pattern has been}%
\typeout{** loaded for the language `#1'. Using the pattern for}%
\typeout{** the default language instead.}%
\else
\language=\csname l@#1\endcsname
\fi
#2}}
\providecommand{\BIBdecl}{\relax}
\BIBdecl

\bibitem{kroemer2019review}
O.~Kroemer, S.~Niekum, and G.~Konidaris, ``A review of robot learning for
  manipulation: Challenges, representations, and algorithms,'' \emph{arXiv
  preprint arXiv:1907.03146}, 2019.

\bibitem{kober2013reinforcement}
J.~Kober, J.~A. Bagnell, and J.~Peters, ``Reinforcement learning in robotics: A
  survey,'' \emph{The International Journal of Robotics Research}, vol.~32,
  no.~11, pp. 1238--1274, 2013.

\bibitem{Argall2009}
B.~D. Argall, S.~Chernova, M.~Veloso, and B.~Browning, ``A survey of robot
  learning from demonstration,'' \emph{Rob. and Auto. Sys.}, vol.~57, no.~5,
  pp. 469--483, 2009.

\bibitem{dulac2019challenges}
G.~Dulac-Arnold, D.~Mankowitz, and T.~Hester, ``Challenges of real-world
  reinforcement learning,'' \emph{arXiv preprint arXiv:1904.12901}, 2019.

\bibitem{hester2018deep}
T.~Hester, M.~Vecerik, O.~Pietquin, M.~Lanctot, T.~Schaul, B.~Piot, D.~Horgan,
  J.~Quan, A.~Sendonaris, I.~Osband \emph{et~al.}, ``Deep q-learning from
  demonstrations,'' in \emph{Thirty-Second AAAI Conference on Artificial
  Intelligence}, 2018.

\bibitem{vecerik2019practical}
M.~Vecerik, O.~Sushkov, D.~Barker, T.~Roth{\"o}rl, T.~Hester, and J.~Scholz,
  ``A practical approach to insertion with variable socket position using deep
  reinforcement learning,'' in \emph{2019 International Conference on Robotics
  and Automation (ICRA)}.\hskip 1em plus 0.5em minus 0.4em\relax IEEE, 2019,
  pp. 754--760.

\bibitem{rajeswaran2018learning}
A.~Rajeswaran, V.~Kumar, A.~Gupta, G.~Vezzani, J.~Schulman, E.~Todorov, and
  S.~Levine, ``Learning complex dexterous manipulation with deep reinforcement
  learning and demonstrations,'' 2018.

\bibitem{barto2002}
A.~G. Barto and T.~G. Dietterich, \emph{Reinforcement Learning and Its
  Relationship to Supervised Learning}.\hskip 1em plus 0.5em minus 0.4em\relax
  John Wiley \& Sons, Ltd, ch.~2, pp. 45--63.

\bibitem{schmidhuber2020reinforcement}
J.~Schmidhuber, ``Reinforcement learning upside down: Don't predict rewards --
  just map them to actions,'' 2020.

\bibitem{srivastava2019training}
R.~K. Srivastava, P.~Shyam, F.~Mutz, W.~Jaśkowski, and J.~Schmidhuber,
  ``Training agents using upside-down reinforcement learning,'' 2019.

\bibitem{NP}
M.~Garnelo, J.~Schwarz, D.~Rosenbaum, F.~Viola, D.~J. Rezende, S.~Eslami, and
  Y.~Teh, ``Neural processes,'' \emph{ArXiv}, vol. abs/1807.01622, 2018.

\bibitem{nolfi2000evolutionary}
S.~Nolfi, D.~Floreano, and D.~D. Floreano, \emph{Evolutionary robotics: The
  biology, intelligence, and technology of self-organizing machines}.\hskip 1em
  plus 0.5em minus 0.4em\relax MIT press, 2000.

\bibitem{Blei_2017}
\BIBentryALTinterwordspacing
D.~M. Blei, A.~Kucukelbir, and J.~D. McAuliffe, ``Variational inference: A
  review for statisticians,'' \emph{Journal of the American Statistical
  Association}, vol. 112, no. 518, p. 859–877, Apr 2017. [Online]. Available:
  \url{http://dx.doi.org/10.1080/01621459.2017.1285773}
\BIBentrySTDinterwordspacing

\bibitem{Seker2019}
M.~Y. Seker, M.~Imre, J.~Piater, and E.~Ugur, ``Conditional neural movement
  primitives,'' in \emph{Proceedings of Robotics: Science and Systems (RSS)},
  Freiburgim, Germany, June 2019.

\bibitem{Akbulut2020}
T.~Akbulut, E.~Oztop, Y.~Seker, H.~Xue, A.~Tekden, and E.~Ugur, ``Acnmp:
  Flexible skill formation with learning from demonstration and reinforcement
  learning via representation sharing,'' in \emph{Conference on Robot Learning
  (CoRL)}, 2020.

\bibitem{peters2010relative}
J.~Peters, K.~M{\"u}lling, and Y.~Altun, ``Relative entropy policy search,'' in
  \emph{Twenty-Fourth AAAI Conference on Artificial Intelligence},
  vol.~10.\hskip 1em plus 0.5em minus 0.4em\relax Atlanta, 2010, pp.
  1607--1612.

\bibitem{colome2017dual}
A.~Colome and C.~Torras, ``Dual reps: A generalization of relative entropy
  policy search exploiting bad experiences,'' \emph{IEEE Transactions on
  Robotics}, vol.~33, no.~4, pp. 978--985, 2017.

\bibitem{ijspeert2013dynamical}
A.~J. Ijspeert, J.~Nakanishi, H.~Hoffmann, P.~Pastor, and S.~Schaal,
  ``Dynamical movement primitives: learning attractor models for motor
  behaviors,'' \emph{Neural computation}, vol.~25, no.~2, pp. 328--373, 2013.

\bibitem{Calinon2016}
S.~Calinon, ``A tutorial on task-parameterized movement learning and
  retrieval,'' \emph{Intelligent Service Robotics}, vol.~9, no.~1, pp. 1--29,
  2016.

\bibitem{Girgin2018}
H.~Girgin and E.~Ugur, ``Associative skill memory models,'' in \emph{{IROS}},
  2018, pp. 6043--6048.

\bibitem{Ugur2020}
E.~Ugur and H.~Girgin, ``Compliant parametric dynamic movement primitives,''
  \emph{Robotica}, vol.~38, no.~3, pp. 457--474, 2020.

\bibitem{Paraschos2013}
A.~Paraschos, C.~Daniel, J.~Peters, and G.~Neumann, ``Probabilistic movement
  primitives,'' in \emph{{NIPS}}, 2013, pp. 2616--2624.

\bibitem{CNP}
M.~Garnelo, D.~Rosenbaum, C.~Maddison, T.~Ramalho, D.~Saxton, M.~Shanahan,
  Y.~W. Teh, D.~Rezende, and S.~M.~A. Eslami, ``Conditional neural processes,''
  in \emph{{ICML}}, 1704-1713 2018.

\bibitem{Chen2016}
N.~Chen, M.~Karl, and P.~van~der Smagt, ``Dynamic movement primitives in latent
  space of time-dependent variational autoencoders,'' 11 2016.

\bibitem{Noseworthy2020}
M.~Noseworthy, R.~Paul, S.~Roy, D.~Park, and N.~Roy, ``Task-conditioned
  variational autoencoders for learning movement primitives,'' ser. Proceedings
  of Machine Learning Research, L.~P. Kaelbling, D.~Kragic, and K.~Sugiura,
  Eds., vol. 100.\hskip 1em plus 0.5em minus 0.4em\relax PMLR, 2020, pp.
  933--944.

\bibitem{Osa_2020}
T.~Osa and S.~Ikemoto, ``Goal-conditioned variational autoencoder trajectory
  primitives with continuous and discrete latent codes,'' \emph{SN Computer
  Science}, vol.~1, no.~5, 2020.

\bibitem{karl2017deep}
M.~Karl, M.~Soelch, J.~Bayer, and P.~van~der Smagt, ``Deep variational bayes
  filters: Unsupervised learning of state space models from raw data,'' 2017.

\bibitem{oord2016wavenet}
A.~van~den Oord, S.~Dieleman, H.~Zen, K.~Simonyan, O.~Vinyals, A.~Graves,
  N.~Kalchbrenner, A.~Senior, and K.~Kavukcuoglu, ``Wavenet: A generative model
  for raw audio,'' 2016.

\bibitem{Stark2019}
S.~Stark, J.~Peters, and E.~Rueckert, ``Experience reuse with probabilistic
  movement primitives,'' in \emph{Proceedings of the IEEE/RSJ International
  Conference on Intelligent Robots and Systems (IROS)}, 2019.

\bibitem{Ewerton2019}
M.~Ewerton, O.~Arenz, G.~Maeda, D.~Koert, Z.~Kolev, M.~Takahashi, and
  J.~Peters, ``Learning trajectory distributions for assisted teleoperation and
  path planning,'' \emph{Frontiers in Robotics and AI}, vol.~6, p.~89, 2019.

\bibitem{salimans2017evolution}
T.~Salimans, J.~Ho, X.~Chen, S.~Sidor, and I.~Sutskever, ``Evolution strategies
  as a scalable alternative to reinforcement learning,'' \emph{arXiv preprint
  arXiv:1703.03864}, 2017.

\bibitem{such2017deep}
F.~P. Such, V.~Madhavan, E.~Conti, J.~Lehman, K.~O. Stanley, and J.~Clune,
  ``Deep neuroevolution: Genetic algorithms are a competitive alternative for
  training deep neural networks for reinforcement learning,'' \emph{arXiv
  preprint arXiv:1712.06567}, 2017.

\bibitem{chang2018genetic}
S.~Chang, J.~Yang, J.~Choi, and N.~Kwak, ``Genetic-gated networks for deep
  reinforcement learning,'' in \emph{Advances in Neural Information Processing
  Systems}, 2018, pp. 1747--1756.

\bibitem{gangwani2017policy}
T.~Gangwani and J.~Peng, ``Policy optimization by genetic distillation,''
  \emph{arXiv preprint arXiv:1711.01012}, 2017.

\bibitem{wang2020instance}
Z.~Wang, C.~Chen, and D.~Dong, ``Instance weighted incremental evolution
  strategies for reinforcement learning in dynamic environments,'' \emph{arXiv
  preprint arXiv:2010.04605}, 2020.

\bibitem{conti2018improving}
E.~Conti, V.~Madhavan, F.~P. Such, J.~Lehman, K.~Stanley, and J.~Clune,
  ``Improving exploration in evolution strategies for deep reinforcement
  learning via a population of novelty-seeking agents,'' in \emph{Advances in
  neural information processing systems}, 2018, pp. 5027--5038.

\bibitem{lehman2011novelty}
J.~Lehman and K.~O. Stanley, ``Novelty search and the problem with
  objectives,'' in \emph{Genetic programming theory and practice IX}.\hskip 1em
  plus 0.5em minus 0.4em\relax Springer, 2011, pp. 37--56.

\bibitem{khadka2018evolution}
S.~Khadka and K.~Tumer, ``Evolution-guided policy gradient in reinforcement
  learning,'' in \emph{Advances in Neural Information Processing Systems},
  2018, pp. 1188--1200.

\bibitem{bodnar2019proximal}
C.~Bodnar, B.~Day, and P.~Li{\'o}, ``Proximal distilled evolutionary
  reinforcement learning,'' \emph{arXiv preprint arXiv:1906.09807}, 2019.

\bibitem{pourchot2018cem}
A.~Pourchot and O.~Sigaud, ``Cem-rl: Combining evolutionary and gradient-based
  methods for policy search,'' \emph{arXiv preprint arXiv:1810.01222}, 2018.

\bibitem{Higgins2017}
I.~Higgins, L.~Matthey, A.~Pal, C.~Burgess, X.~Glorot, M.~Botvinick,
  S.~Mohamed, and A.~Lerchner, ``beta-vae: Learning basic visual concepts with
  a constrained variational framework,'' in \emph{ICLR}, 2017.

\bibitem{Sohn2015}
K.~Sohn, H.~Lee, and X.~Yan, ``Learning structured output representation using
  deep conditional generative models,'' in \emph{Advances in Neural Information
  Processing Systems 28}, C.~Cortes, N.~D. Lawrence, D.~D. Lee, M.~Sugiyama,
  and R.~Garnett, Eds.\hskip 1em plus 0.5em minus 0.4em\relax Curran
  Associates, Inc., 2015, pp. 3483--3491.

\bibitem{Diederik2019}
\BIBentryALTinterwordspacing
D.~P. Kingma and M.~Welling, ``An introduction to variational autoencoders,''
  \emph{CoRR}, vol. abs/1906.02691, 2019. [Online]. Available:
  \url{http://arxiv.org/abs/1906.02691}
\BIBentrySTDinterwordspacing

\bibitem{doersch2016tutorial}
C.~Doersch, ``Tutorial on variational autoencoders,'' 2016.

\end{thebibliography}

\end{document}